\documentclass[10pt,twocolumn,letterpaper]{article}

\usepackage[pagenumbers]{cvpr} 
\usepackage{multirow}
\usepackage{graphicx}
\usepackage{amsmath}
\usepackage{dcolumn} 
\usepackage{tikz}

\usetikzlibrary{positioning}

\definecolor{cvprblue}{rgb}{0.21,0.49,0.74}
\usepackage[pagebackref,breaklinks,colorlinks,citecolor=cvprblue]{hyperref}

\def\paperID{*****} 
\def\confName{CVPR}
\def\confYear{2025}

\title{The Context of Crash Occurrence: A Complexity-Infused Approach Integrating Semantic, Contextual, and Kinematic Features}

\author{Meng Wang\\
University of Massachusetts Amherst\\
{\tt\small mwang0@umass.edu}
\and 
Zach Noonan\\
Massachusetts Institute of Technology\\
{\tt\small tznoonan@mit.edu}
\and
Pnina Gershon\\
Massachusetts Institute of Technology\\
{\tt\small pgershon@mit.edu}
\and
Bruce Mehler\\
Massachusetts Institute of Technology\\
{\tt\small bmehler@mit.edu}
\and
Bryan Reimer\\
Massachusetts Institute of Technology\\
{\tt\small reimer@mit.edu}
\and
Shannon C. Roberts\\
University of Massachusetts Amherst\\
{\tt\small scroberts@umass.edu}
}

\begin{document}
\maketitle
\begin{abstract}
Understanding the context of crash occurrence in complex driving environments is essential for improving traffic safety and advancing automated driving. Previous studies have used statistical models and deep learning to predict crashes based on semantic, contextual, or vehicle kinematic features, but none have examined the combined influence of these factors. In this study, we term the integration of these features ``roadway complexity''. This paper introduces a two-stage framework that integrates roadway complexity features for crash prediction. In the first stage, an encoder extracts hidden contextual information from these features, generating complexity-infused features. The second stage uses both original and complexity-infused features to predict crash likelihood, achieving an accuracy of 87.98\% with original features alone and 90.15\% with the added complexity-infused features. Ablation studies confirm that a combination of semantic, kinematic, and contextual features yields the best results, which emphasize their role in capturing roadway complexity. Additionally, complexity index annotations generated by the Large Language Model outperform those by Amazon Mechanical Turk, highlighting the potential of AI based tools for accurate, scalable crash prediction systems.
\end{abstract}    
\section{Introduction}
\label{sec:intro}

\begin{figure*}[t]
  \centering
   \includegraphics[width=1\linewidth]{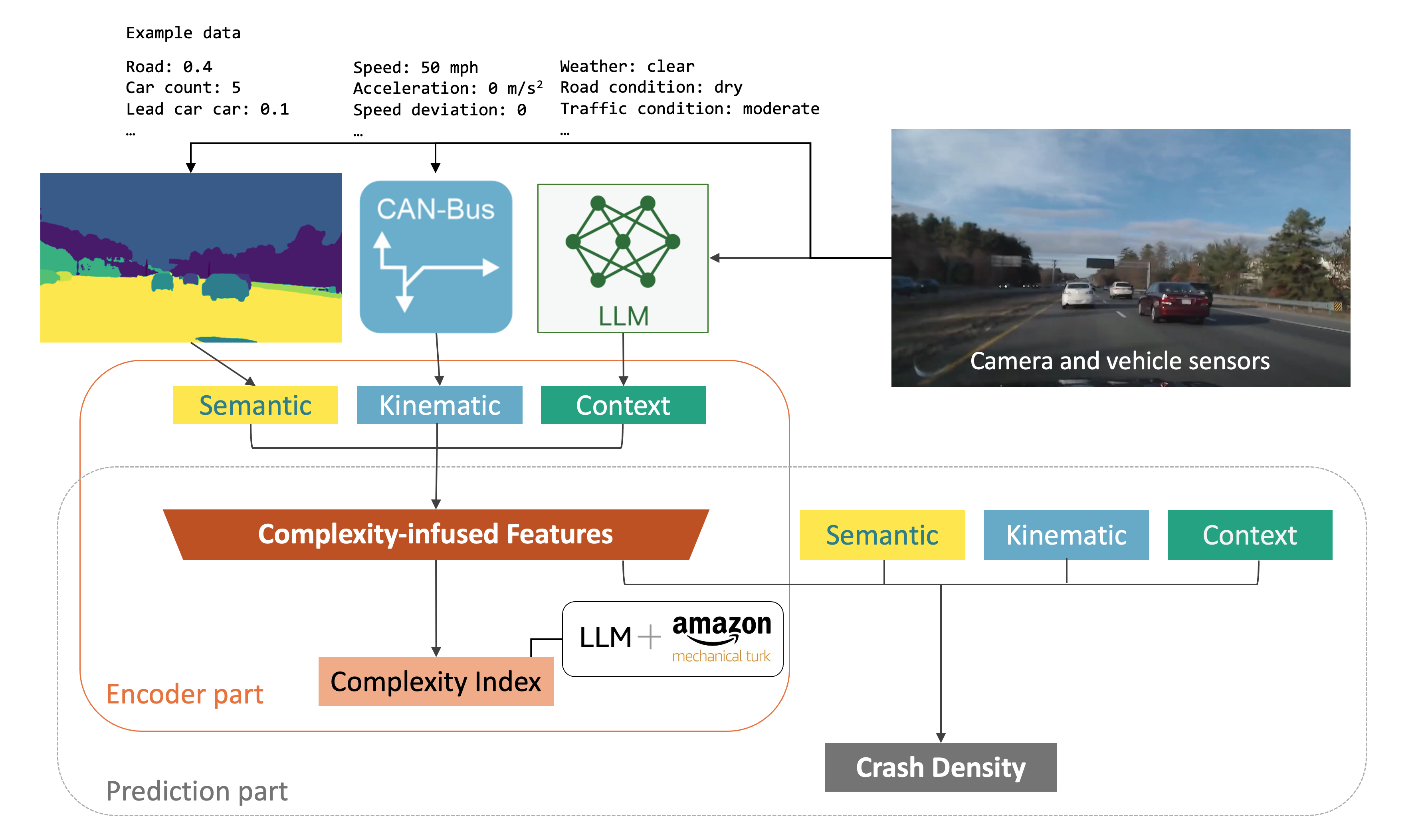}
   \caption{The model structure. The model takes raw images and CAN-Bus signals as inputs to generate semantic, contextual, and kinematic features, which are then used to investigate their relationship with crash density estimates, serving as the model’s output. It consists of an encoder that learns hidden features from the semantic, kinematic, and contextual data, which are infused with the complexity index. The prediction model then utilizes all the available features, including the complexity-infused features, to predict the crash density and rates. Example data is shown above each feature source.}
   \label{fig:model}
\end{figure*}

Understanding factors contributing to crashes is essential for improving traffic safety and advancing automated vehicle design. A 2015 NHTSA report~\cite{singh2015critical} found that 94\% of 5,470 crashes from 2005 to 2007 were due to human errors like inattention, distractions, excessive speed, and misjudgment.

Research has sought to predict crash frequency and rates based on environmental factors and driver behavior. Statistical models have linked crash occurrences to variables such as speed, traffic volume, weather, and road design~\cite{wang2013effect, rashidi2022modeling, xu2018association, hammad2019environmental, xu2013identifying}, while recent studies have used deep learning to analyze real-time images and sensor data from vehicles and infrastructure to detect hazards~\cite{karim2021system, hu2023image}. Though promising, these models rely on either imagery or behavioral data, without integrating both to predict crash occurrences or density.

In this paper, we define \textit{roadway scene complexity} as a combination of semantic information (\eg, number of objects) and contextual variation (\eg, road curvature, roadway type, weather). Scene complexity affects driver behavior, with factors like object density and road type impacting drivers’ situational awareness~\cite{park2022impact, asteriou2024characterizes}. Visual and environmental complexity, such as high traffic volumes or urban versus rural settings, also influences cognitive demands on drivers~\cite{kaber2012effects, hao2007effect, mohan2017urban, mukherjee2019impact}.

Driving behavior, such as speed adjustments in response to poor visibility or narrow lanes, is also influenced by scene complexity. Speed and acceleration patterns adjust based on obstacles and conditions~\cite{khan2020evaluating}. Integrating behavior data with scene information deepens our understanding of driver interactions with their environment, improving crash risk modeling. Thus, we define \textit{roadway complexity} as the combination of scene complexity and driving behavior, where driving behavior is represented by vehicle kinematic features.

Extracting hidden context from this combined data is essential. Previous studies have shown that fusing situational and memory-based features~\cite{zhu2021improving}, as well as road networks and motion history data~\cite{mu2024most}, enhances situation awareness and motion prediction, respectively. Building on this, we incorporate feature fusion to capture both explicit and implicit features of roadway scenes.

To address the challenges of (1) studying roadway complexity more holistically by incorporating both imagery and behavioral data, and (2) investigating the direct and indirect relationships between roadway complexity and crash density and rates, we propose a two-stage fusion prediction model framework.  As illustrated in Fig.~\ref{fig:model}, the framework learns hidden features from the fusion of the semantic representation of the driving scene, vehicle kinematic features, and contextual features. The framework uses a combination of these features to form a comprehensive representation of roadway complexity and its association with crash density. Our model, which is trained on a naturalistic dataset and historical crash data, consists of an encoder that captures the hidden context of the roadway complexity and a prediction module that uses these features to investigate the relationship between the crash density. The contributions of this paper are:

\begin{enumerate}
    \item We take a holistic approach to studying roadway complexity by considering both imagery information, \ie, specifically, semantic representation of the scene and contextual features, and vehicle kinematic data, and we empirically demonstrate the necessity of incorporating all these elements.
    \item We argue that roadway complexity can be directly and indirectly linked to crash density and rates. We introduce a two-stage framework that extracts hidden contextual features of roadway scene complexity by fusing semantic, kinematic, and contextual data, and we investigate the predictive power of both direct and hidden features to identify features associated with crash density and rates.
    \item We compare the complexity index annotations from Amazon Mechanical Turk and the Large Language Model (LLM) in terms of their capability to predict crash density and find that LLM-generated annotations consistently exhibited better predictive performance. This can enhance the development of real-time crash prediction systems and inspire the integration of automated annotation tools for improved accuracy and scalability.
\end{enumerate}

\section{Related Work}
\label{sec:related}

There has been extensive research on understanding roadway scenes using both Convolutional Neural Network (CNN)-based and transformer-based methods. Panoptic segmentation, which unifies semantic and instance segmentation, offers a holistic understanding by labeling all pixels and differentiating object instances, making it especially useful in structured environments like roadways~\cite{Kirillov_2019_CVPR}. The task has been successfully applied to driving scene datasets, such as CityScapes~\cite{Cordts2016Cityscapes}, which provides high-resolution images for semantic urban scene understanding. Recent advancements in transformer-based models have further improved performance in scene understanding. Among these, OneFormer~\cite{jain2023oneformer} has achieved state-of-the-art panoptic quality scores~\cite{Kirillov_2019_CVPR} while maintaining computational efficiency, making it suitable for real-time scene understanding in automated driving. Given its strong performance, the OneFormer model was utilized in this study to extract scene-level visual features from the driving environments.

Understanding the driving context is essential for both human drivers and automated systems, as it involves not only the perception of roadway scene elements but also can help interpret the interactions between these elements. Recent advancements in LLMs have demonstrated their potential to go beyond pixel-level interpretation, incorporating higher-level reasoning into driving tasks. LLMs have been applied to extract contextual information from driving environments by mapping raw sensory data (\eg, images, LiDAR) to higher-level contextual descriptions~\cite{sural2024contextvlm}. This is particularly useful for driving research, where understanding the ``intent'' of other road users, predicting possible hazards, or inferring rules of the road from visual cues may significantly enhance safety and driver performance~\cite{cui2023drivellm, yildirim2024highwayllm}. The application of LLMs indicates a leap forward as they can go beyond object detection to understand the context of the driving scenarios.

\section{Methods}
\label{sec:method}

In this section, we outline the dataset that was used in the study and the methodology for generating the features used in the crash density prediction model. We detail the architecture of the complexity-infused encoder and describe the structure of the final prediction model.

\subsection{Dataset Overview}
This study utilizes a subset dataset derived from the MIT-AVT naturalistic driving studies~\cite{fridman2019advanced}. The dataset includes a variety of multimodal data sources, such as 20-second raw video clips from the forward-facing camera, 30 Hz CAN bus data, 30 Hz GPS data, and contextual metadata features. The contextual features are sourced from the study by~\citet{ding2020avt}.

For the purposes of this research, 500 video clips were selected from the dataset developed from~\citet{ding2020avt}, categorized as follows: 100 highway scenarios, 100 rural scenarios, 100 urban scenarios, 75 bridge scenarios, 75 overpass scenarios, and 75 crash hotspot scenarios. The methodology for generating crash hotspot data is detailed in Section~\ref{crash}.

From each selected video clip, frames were extracted based on a fixed-distance sampling method, with one frame captured for every 20 meters traveled. The extraction began with the first frame of each clip and continued until the distance to the subsequent frame was less than 20 meters. Overall, there were 10,407 frames extracted.

\subsection{Feature Generation}
For each image frame, three sets of features were generated: semantic features derived from a computer vision algorithm, vehicle kinematic features extracted from the CAN bus data, and contextual features generated from the LLM. 

\subsubsection{Semantic Features}

\begin{figure}[ht!]
  \centering
   \includegraphics[width=1\linewidth]{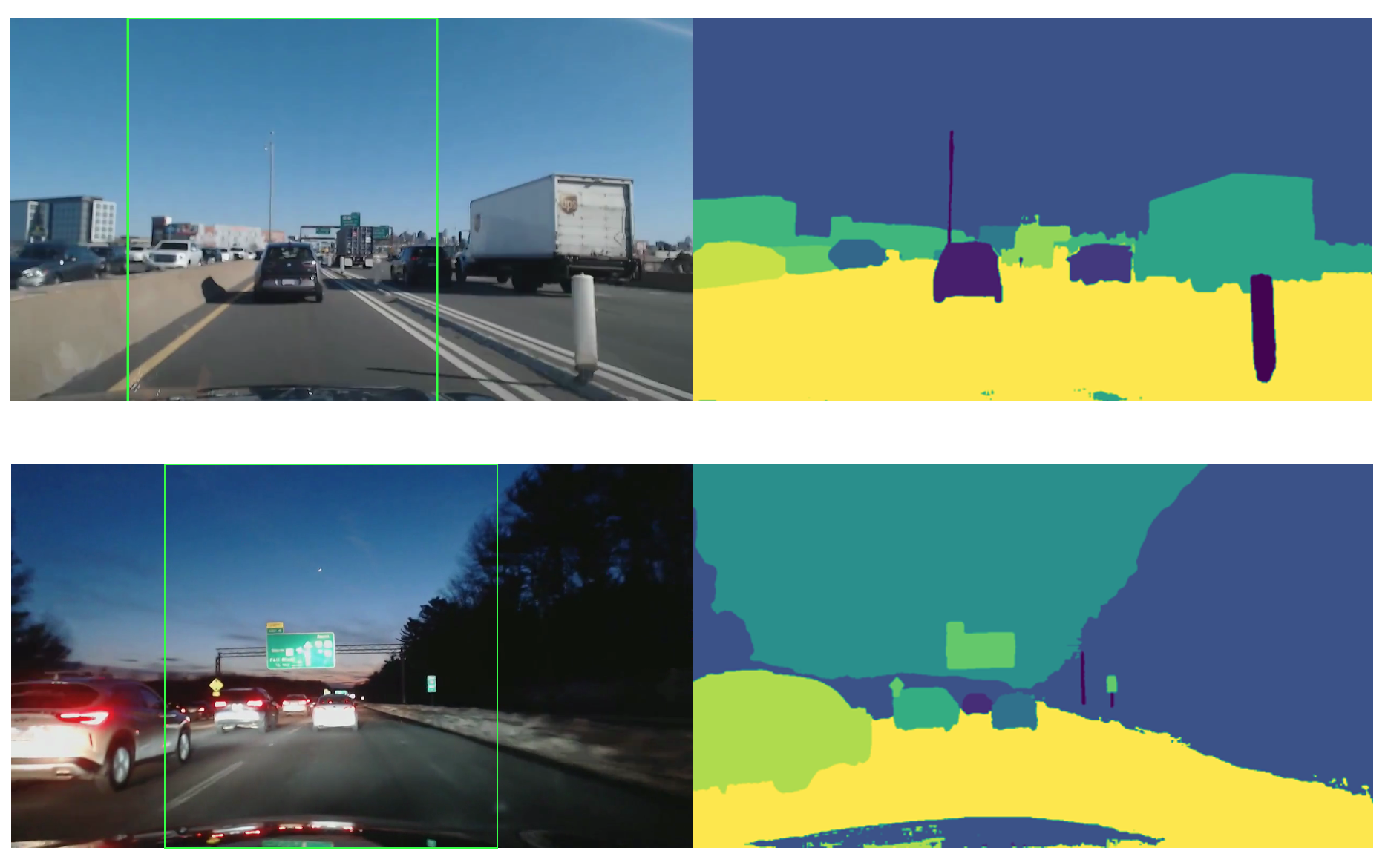}
   \caption{The raw roadway scene image and OneFormer algorithm output. The lead-car region is highlighted in a green box. }
   \label{fig:example}
\end{figure}

The semantic features were generated using the OneFormer algorithm~\cite{jain2023oneformer} via Hugging Face’s API\footnote{https://huggingface.co/docs/transformers/en/model\_doc/oneformer}. The model outputs pixel-level semantic classifications for various objects such as cars, pedestrians, bicycles, roads, traffic signs, sidewalks, buildings, vegetation, and sky. To better understand the effect of complexity on driver behavior, a lead-car region was defined~\cite{yang2023takeover}. An example of the OneFormer algorithm’s output is provided in Fig.~\ref{fig:example}.

For each image, the percentage of pixels corresponding to each class relative to the total frame size was calculated. Similarly, the percentage of pixels for each class within the lead-car region was computed. Additionally, the number of cars, pedestrians, buses, bicycles, and motorcycles was counted in both the full frame and the lead-car regions. This process resulted in 50 initial semantic features. After closer examination, features with minimal variability (more than $90\%$ being $0$) were removed, reducing the final set to 17 semantic features. The final set of features is coded as:

\begin{enumerate}
\item Full frame (10): `car\_count', `road', `vegetation', `sky', `terrain', `car', `sidewalk', `building', `traffic\_light', `person'. 
\item Lead car region (7): `lead\_car\_traffic\_sign', `lead\_car\_road', `lead\_car\_vegetation', `lead\_car\_sky', `lead\_car\_car', `lead\_car\_fence', `lead\_car\_car\_count'.
\end{enumerate}

\subsubsection{Vehicle Kinematic Features}
The vehicle kinematic features were extracted from the CAN bus data. For each frame and the corresponding 20-meter segment, the following 9 features were computed: current speed, mean speed, standard deviation of speed, mean longitudinal acceleration, standard deviation of longitudinal acceleration, minimum longitudinal acceleration, maximum longitudinal acceleration, raw deviation from the speed limit, and normalized deviation from the speed limit. The raw and normalized deviation from the speed limit are defined as follows:
\begin{equation}
  SpeedDev = Speed_{mean} - SpeedLimit
\end{equation}
\begin{equation}
  {SpeedDev}_{normalized} = \frac{Speed_{mean} - SpeedLimit}{SpeedLimit}
\end{equation}

To obtain the speed limit information for each road segment, Microsoft Azure’s Get Search Address Reverse API \footnote{https://learn.microsoft.com/en-us/rest/api/maps/search/get-search-address-reverse?view=rest-maps-1.0\&tabs=HTTP} was utilized. 


\subsubsection{Contextual Features} \label{cont}
\begin{figure}[ht!]
  \centering
   \includegraphics[width=1\linewidth]{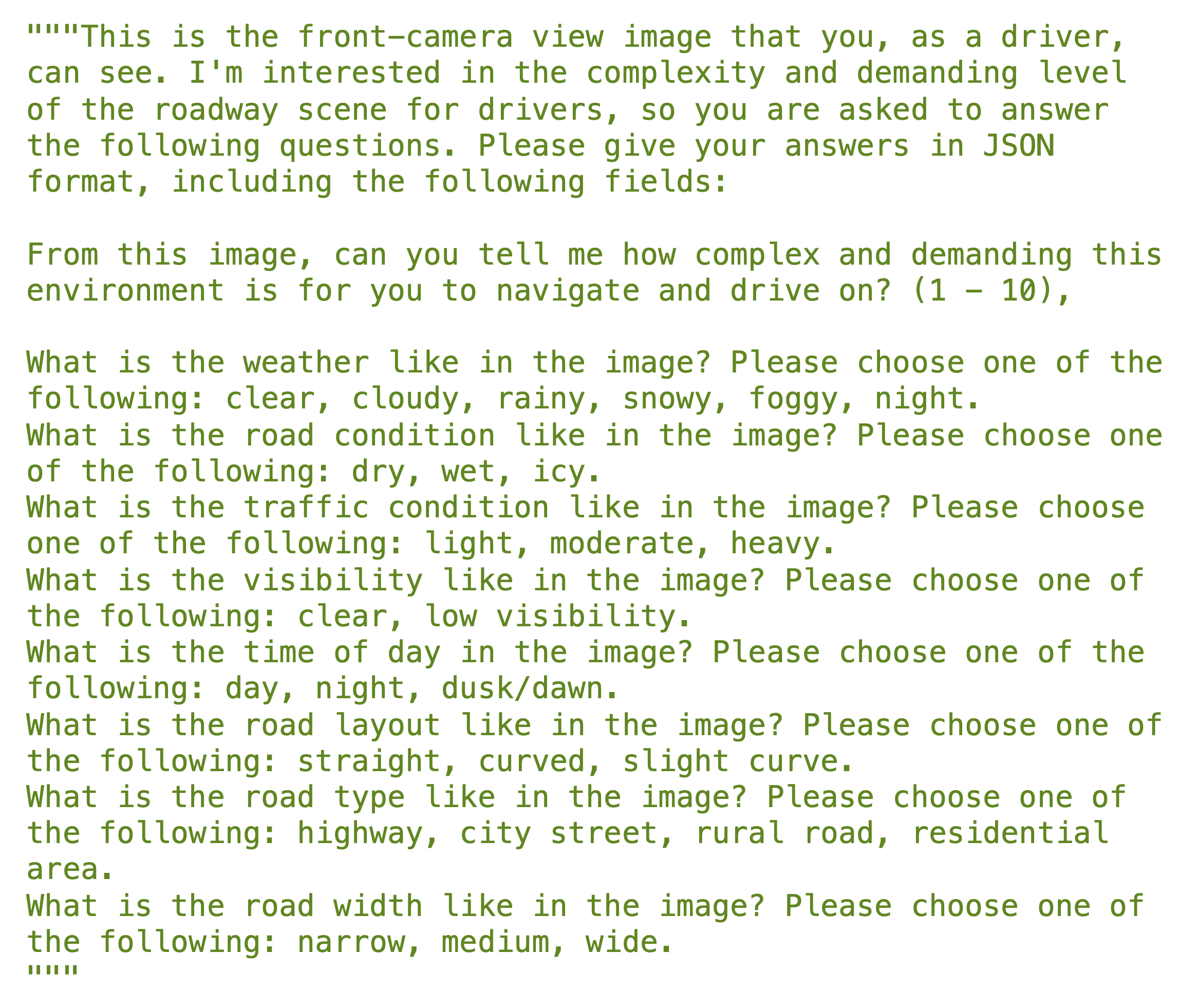}
   \caption{The prompt used in collecting contextual features with GPT-4o model.}
   \label{fig:prompt}
\end{figure}

The contextual features include road characteristics, which were generated using the GPT model~\cite{achiam2023gpt}. The implementation details can be found in Appendix~\ref{sup:LLM}. The prompt used in this step is illustrated in Fig.~\ref{fig:prompt}. In the prompt, several questions were asked to gather contextual road characteristics, including information on weather conditions, road conditions, traffic conditions, visibility levels, time of day, road layout, road type, and lane width. Initially, each question was presented in an open-ended format to generate a pool of possible answers. The prompt was then refined based on these responses, providing predefined options to ensure consistency across the answers. During the prompting process, the GPT model was asked to generate a single output option for each question in the prompt, which was formatted in JSON. The prompt was run three times on each image to ensure data quality. For each question, the final answer was determined by selecting the response that was agreed upon by the majority of the three outputs. 

During analysis, these contextual features were one-hot encoded to convert them into numerical values for integration into the model. In total, this process generated 19 contextual features. As can be seen in Fig.~\ref{fig:prompt}, there were 5 features for weather, 2 for road condition, 2 for traffic condition, 1 for visibility, 2 for time of day, 2 for road layout, 3 for road type, and 2 for lane width.

\subsection{Ground Truth Generation} \label{crash}
There were two ground truths in this study. As illustrated in Figure~\ref{fig:model}, for the encoder, the complexity index was collected to quantify the overall complexity level in a given roadway scene. For the prediction, the crash density value was computed and serves as the ground truth for estimating crash rates.

\subsubsection{Complexity Index} \label{comp}
The complexity index was generated from two sources: AI and humans. For AI, the GPT model was used along with the contextual feature generation process, as shown in Fig.~\ref{fig:prompt}. In this approach, the model generated a complexity score on a scale from 0 to 10 to describe the complexity and demand level of the roadway scenes.

The human-generated complexity indices relied on Amazon Mechanical Turk (MTurk) for annotations. The task was designed to assess the complexity level of roadway scenes. Workers were shown image frames and asked to rate the complexity of each scene on a scale from 1 to 10. Only workers with a high approval rating, at least 500 completed tasks, and residing in the US were selected. A pilot study was conducted with 500 images, where 10 workers annotated the same image. The results showed a relatively high level of agreement among workers, with an average standard deviation of $1.74$ per image and a maximum standard deviation of less than $3$ for all images. Based on this, each scene for the study was annotated by 3 workers, and the final complexity score was determined by averaging their responses. Further details on the comparison between AI and human-generated annotations can be found in Appendix~\ref{sup:annoComp}.

\subsubsection{Crash Density}
To generate the crash density, historical crash data was obtained from the Massachusetts Department of Transportation’s IMPACT app \footnote{https://apps.impact.dot.state.ma.us/cdp/home}. GPS data from crashes over a 5-year period (2018 to 2022) was aggregated. A Kernel Density Estimation (KDE) method was then applied to this GPS data to create a continuous scale representing crash density. The Kernel Density Estimation is formulated as follows:

\begin{equation}
  Density = \frac{1}{(radius)^2}\sum[\frac{3}{\pi} \cdot (1 - (\frac{{dist}_i}{radius})^2)^2],
\end{equation}

\noindent for ${dist}_i < radius$, where $i = 1, \dots, n$ are the input points. Only the points within the $radius$ distance of the $(x, y)$ location will be included. ${dist}_i$ is the distance between point $i$ and the $(x, y)$ location. In this study, a radius of 1000 meters was used.

\begin{figure*}[ht!]
  \centering
   \includegraphics[width=1\linewidth]{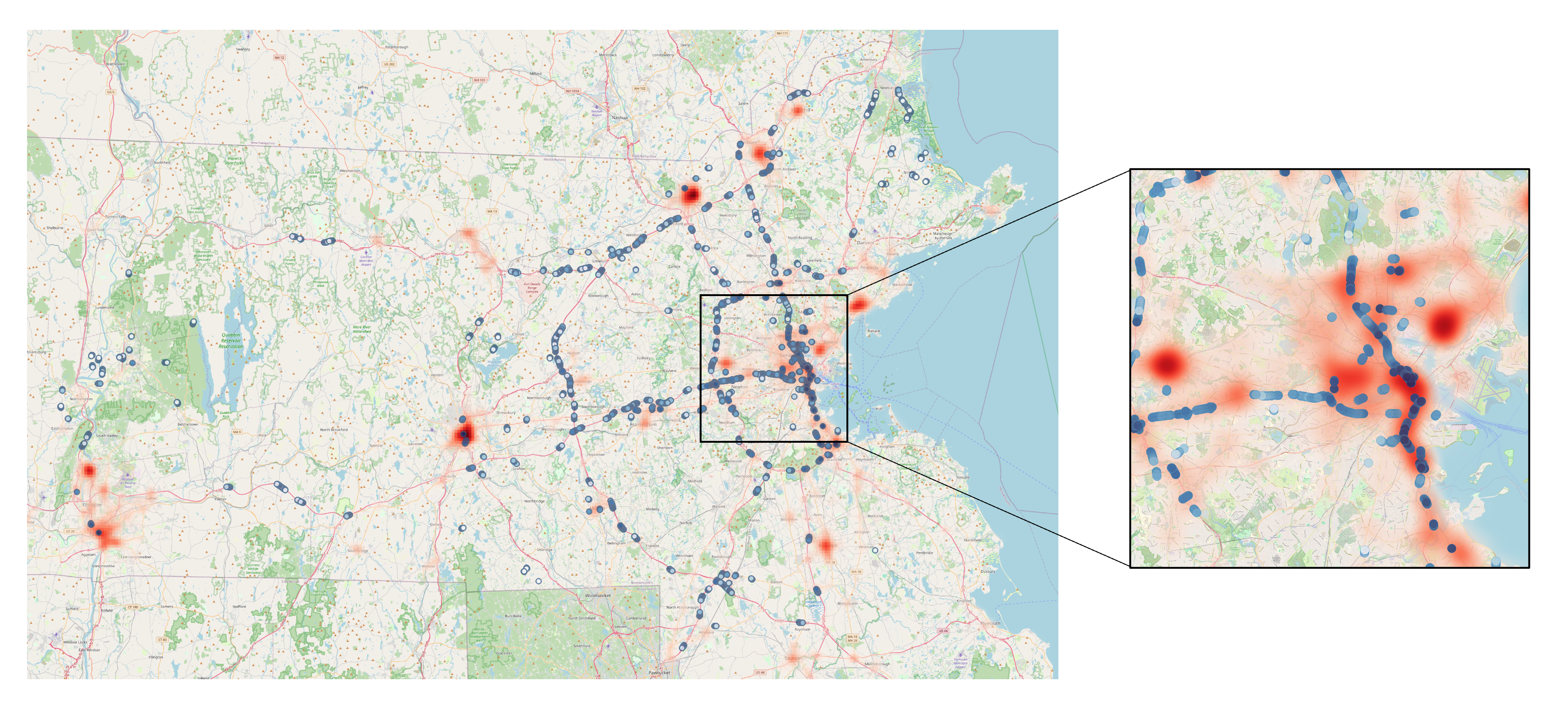}
   \caption{Crash density heatmap (2018-2022) in Massachusetts, displayed in red, where darker colors indicate a higher crash density. Five hundred video clips from the MIT-AVT dataset are marked in blue.}
   \label{fig:heatmap}
\end{figure*}

After applying the KDE method, each location was assigned a corresponding crash density value. The GPS data from the MIT-AVT video clips was then mapped onto this heatmap so that each 20-meter segment was associated with an estimated crash density. The resulting crash density heatmap and the 500 video clips selected from the MIT-AVT dataset are shown in Fig.~\ref{fig:heatmap}.

\begin{figure}[ht!]
  \centering
   \includegraphics[width=.8\linewidth]{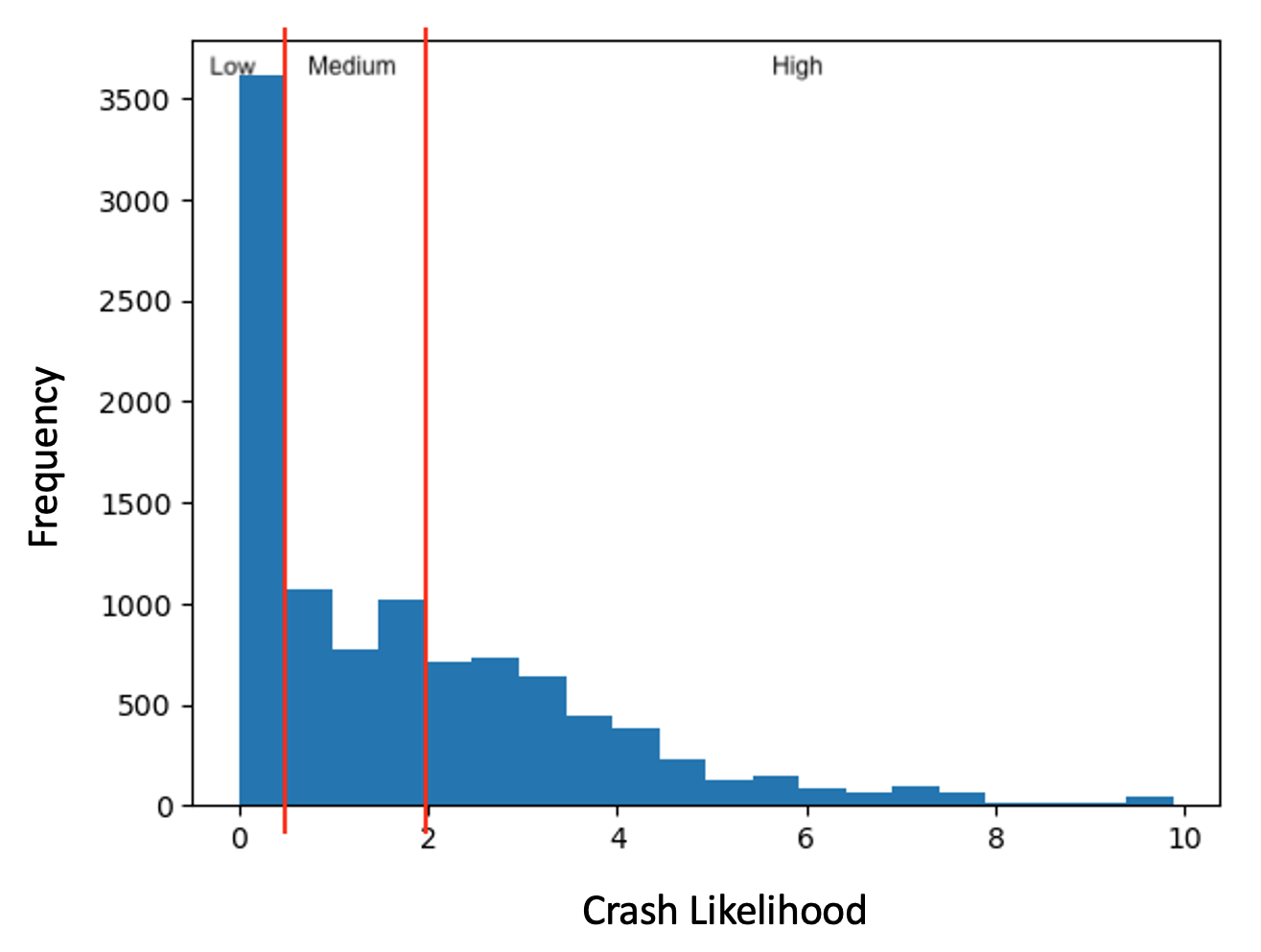}
   \caption{The distribution of crash density value.}
   \label{fig:crash}
\end{figure}

To keep the scale consistent with the Complexity Index, the crash density was normalized and represented on a scale from 0 to 10. The 10,407 frames were categorized into three levels based on the calculated crash density value. As shown in Fig.~\ref{fig:crash}, the crash density distribution was skewed and unbalanced. There was a noticeable trend where crash density ranges between 0 and 0.5 are highly frequent, decrease somewhat in the range of 0.5 to 2, and become increasingly rare beyond 2.  Therefore, density values between 0 and 0.5 were categorized as \textit{Low}, those between 0.5 and 2 as \textit{Medium}, and those between 2 and 10 as \textit{High}. This classification resulted in approximately 3,600 images in the low category, 2,800 in the medium category, and 3,700 in the high category.

\subsection{Modeling}
\subsubsection{Complexity-infused Encoder} \label{encoder}
The complexity-infused encoder used a fully connected neural network structure with either 16 or 32 hidden neurons. The input to the network was threefold: (1) the 17 semantic features, (2) the combination of the 17 semantic features and 9 kinematic features, or (3) the combination of all features—17 semantic features, 9 kinematic features, and 19 contextual features. The input variables were normalized to a 0-1 range to ensure consistency across features and improve the stability of the model during training. The output of the network was the complexity index, which was treated as either a continuous or categorical variable for data obtained from the LLM, and as a continuous variable for data obtained from MTurk. The Root Mean Square Error (RMSE) was used as the evaluation metric for the complexity index when treated as a continuous variable, while accuracy was used as the metric when the complexity index was treated as a categorical variable.

After the hidden layer, a ReLU activation function is applied. The features generated after the ReLU activation are referred to as the complexity-infused features. 

\subsubsection{Crash Density Prediction Model} \label{crashDensity}
After generating the complexity-infused features from the encoder, they were used to predict the level of crash density value in combination with the corresponding input feature sets. For example, if the complexity-infused features were trained on only semantic features, the input for the crash prediction model would consist of both the complexity-infused and semantic features. Similarly, if the complexity-infused features were trained on all available features, the input to the crash prediction model would include the complexity-infused, semantic, kinematic, and contextual features.

To train the crash prediction model, various algorithms were tested, including Random Forest (RF), Gradient-Boosted Decision Trees (GBDT), K-Nearest Neighbors (KNN), and fully connected neural networks (NN). The NN model used in this step is a fully connected neural network consisting of seven linear layers. The detailed structure of the NN model can be found in Appendix~\ref{sup:structure}. Similar to the encoder described in Section~\ref{encoder}, the input variables were normalized to a 0-1 range. The dataset was split into 70\% for training and 30\% for testing, and model performance was evaluated using accuracy as the primary metric. To ensure consistency, the dataset was split in the same way as it was for the encoder.

For best-performing tree-based models, the SHAP values~\cite{lundberg2020local2global} were used to analyze the impact of the most influential features associated with crash density.

\section{Experiments and Results}
\label{sec:results}
This section presents the modeling results of the encoder and the crash density prediction models. Section~\ref{sec:encoder} and Section~\ref{sec:main_result} discuss the results obtained using AI-generated annotations, specifically from the GPT model, as the crash density prediction models utilizing AI annotations consistently outperformed those using human annotations. Section~\ref{sec:ablation} provides a comparative analysis of AI annotations versus human annotations, as well as an evaluation of different encoder structures.

\subsection{Complexity-infused Encoder} \label{sec:encoder}
\textbf{Implementation.} The complexity-infused encoder was implemented with different settings, depending on the input feature sets and the type of output variable, \ie, the complexity index. When the complexity index was treated as a continuous variable, the Root Mean Squared Error (RMSE) was used as the loss function. When the index was treated as a categorical variable, Cross Entropy (CE) was chosen as the loss function. Further details of the model training configurations can be found in Appendix~\ref{sup:encoderTrainingDetails}.

\noindent \textbf{Results. }
The model performance of the encoder is presented in Table~\ref{tab:encoder_performance}, showing the model performance on the complexity index obtained from the LLM. As shown in the table, the three model settings exhibited similar performance across different input feature sets. However, several insights can be drawn from the results: first, the model with 32 hidden neurons and a continuous output produced the lowest RMSE overall. Second, models with more comprehensive input feature sets, which incorporated semantic, kinematic, and contextual features, consistently outperformed those with fewer feature sets. This pattern was observed across all three models.

\begin{table}[ht!]
\footnotesize
\begin{tabular}{llcc}
\hline
\textbf{Model Settings }                                                                                                  & \textbf{Input Features}              & \begin{tabular}[c]{@{}c@{}}\textbf{RMSE/CE} \\ \textbf{on Train} $\downarrow$\end{tabular} & \begin{tabular}[c]{@{}c@{}}\textbf{RMSE/CE} \\ \textbf{on Test} $\downarrow$\end{tabular} \\ \hline
\multirow{3}{*}{\textbf{\begin{tabular}[c]{@{}l@{}}32 neurons, \\ 1-d cont. output\end{tabular}}} & \textbf{Semantic}           & \textbf{$1.10$}                                               & \textbf{$1.10$}                                              \\
                                                                                                              & \textbf{Semantic + kinematic} & \textbf{$1.05$}                                               & \textbf{$1.06$}                                              \\
                                                                                                              & \textbf{Sem. + Kin. + Cont.  }                & \textbf{$0.84$}                                               & \textbf{$0.86$}                                              \\ \hline
\multirow{3}{*}{\begin{tabular}[c]{@{}l@{}}16 neurons, \\ 1-d cont. output\end{tabular}}          & Semantic                    & $1.11$                                                        & $1.12$                                                       \\
                                                                                                              & Semantic + kinematic          & $1.08$                                                        & $1.08$                                                       \\
                                                                                                              & Sem. + Kin. + Cont.                           & $0.84$                                                        & $0.85$                                                       \\ \hline
\multirow{3}{*}{\begin{tabular}[c]{@{}l@{}}32 neurons, \\ 10-d cat. output\end{tabular}}        & Semantic                    & $1.33$                                                        & $1.35$                                                       \\
                                                                                                              & Semantic + kinematic          & $1.31$                                                       & $1.34$                                                       \\
                                                                                                              & Sem. + Kin. + Cont.                         & 1$.15$                                                        & $1.18$                                                       \\ \hline
\end{tabular}
\caption{Performance comparison of the complexity-infused encoder models with different input feature sets and hidden neurons. The results are reported in terms of RMSE for continuous outputs and Cross-Entropy for categorical outputs. Downward arrows next to the values indicate that lower values are preferable.}
\label{tab:encoder_performance}
\end{table}

Since the encoder with 32 hidden neurons and a continuous output variable yielded the best results, the 32 hidden features from this model were used as the complexity-infused features in Section~\ref{sec:main_result}.

\subsection{Main Results of Crash Prediction Model} \label{sec:main_result}
\textbf{Implementation.} When training the neural network for the crash prediction model, Cross Entropy was chosen as the loss function during training. The accuracy of the best-performing model is reported in the next section. Further details on the model training configurations can be found in Appendix~\ref{sup:predictionTrainingDetails}.

\begin{table*}[h]
\small
\begin{tabular}{ll|ccc|cc}
\hline
\textbf{Input Features}                                   & \textbf{Model} & \textbf{Baseline} $\uparrow$       & \textbf{+ Comp.-infused} $\uparrow$ & \textbf{Difference}    & \textbf{Comp.-infused alone} $\uparrow$ & \textbf{+ Comp. Index} $\uparrow$ \\ \hline
\multirow{4}{*}{Semantic}                        & RF    & \textbf{73.23} & \textbf{78.35}       & 5.12          & 65.31                    & 74.07              \\
                                                 & GBDT  & 68.42          & 72.49                & 4.07          & 60.97                    & 68.30              \\
                                                 & KNN   & 66.09          & 70.86                & 4.77          & 63.85                    & 67.49              \\
                                                 & NN    & 67.81          & 73.11                & 5.30          & 61.07                    & 68.83              \\ \hline
\multirow{4}{*}{Semantic + kinematic}              & RF    & \textbf{83.56} & \textbf{86.32}       & 2.76          & 67.85                    & 84.07              \\
                                                 & GBDT  & 74.62          & 77.41                & 2.79          & 62.24                    & 74.88              \\
                                                 & KNN   & 74.15          & 75.98                & 1.83          & 67.42                    & 73.53              \\
                                                 & NN    & 75.60          & 77.95                & 2.35          & 66.08                    & 72.45              \\ \hline
\multirow{4}{*}{\begin{tabular}[c]{@{}l@{}}Semantic + kinematic + \\ Contextual\end{tabular}} & RF    & \textbf{87.98} & \textbf{90.15}       & 2.17          & 78.49                    & 87.35              \\
                                                 & GBDT  & 80.13          & 82.24                & 2.11          & 68.03                    & 80.04              \\
                                                 & KNN   & 81.41          & 82.48                & 1.07          & 76.74                    & 81.25              \\
                                                 & NN    & 80.08          & 88.78                & 8.70          & 73.01                    & 84.53              \\ \hline
\end{tabular}%
\caption{Performance comparison of different combinations of feature sets for crash likelihood prediction. The results are reported in terms of accuracy (\%) on the test set. Upward arrows next to the values indicate that higher accuracy is preferable.}
\label{tab:main_results}
\end{table*}

\begin{figure*}[ht!]
  \centering
   \includegraphics[width=1\linewidth]{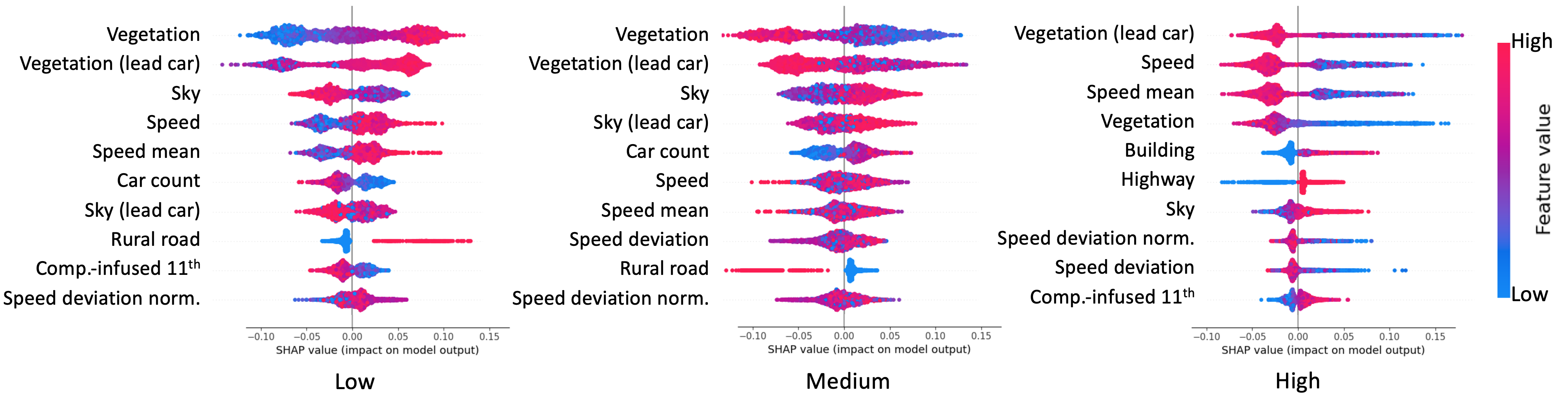}
   \caption{The SHAP values of the 10 most influential features for each class of the best-performing model.}
   \label{fig:shap}
\end{figure*}

\noindent \textbf{Results.}
The performance of the crash prediction model is shown in Table~\ref{tab:main_results}. The baseline model was trained using only the original feature sets, and the complexity-infused features were subsequently added to investigate any improvement in model performance. Additionally, for comparison, the performance of the model trained using only the complexity-infused features was evaluated, as well as the effect of adding the complexity index to the baseline model.

The results indicated that the Random Forest model consistently achieved the best performance across all combinations of input feature sets. Additionally, there was a clear trend of improved model performance as the number of input features increased, with the highest accuracy reaching 87.98\%. Adding the complexity-infused features led to further improvements in prediction performance, with the highest accuracy being 90.15\%, when using semantic, kinematic, and contextual features. This was followed by the neural network model, which achieved an accuracy of $88.78\%$. To assess the significance of these improvements, McNemar’s test~\cite{mcnemar1947note} was conducted for each model, and the results indicated that the improvements were statistically significant. The results suggested that incorporating complexity-infused features, particularly in models using a combination of semantic, kinematic, and contextual features, significantly enhances the accuracy of crash likelihood predictions. 

For the best-performing model, which was trained on all available features, including complexity-infused features, using the Random Forest algorithm, the SHAP values of the 10 most influential features for each class are shown in Fig.~\ref{fig:shap}. The SHAP values highlight the features that most strongly differentiate between each pair of classes. For example, when differentiating the low-density class from the medium-density class, the vegetation area in both the full frame and the lead-car region emerged as the most significant features. Higher values of vegetation area were associated with low crash density areas. 

For the high-density class, speed-related features were the most influential. The results revealed that lower speeds are associated with higher crash-density areas. Regarding contextual features, rural roads were associated with medium crash-density areas compared to low-density areas, while highways were more frequently associated with high crash-density areas. Among the complexity-infused features, at least one was found to be influential in differentiating between low and high crash-density areas.


Using the complexity-infused features alone or adding the 1-dimensional complexity index to the baseline models did not outperform the baseline models, except for the neural network model on semantic, kinematic, and contextual features. This suggests that simply adding the complexity index does not substantially improve the model’s prediction capability. However, the complexity-infused features were found to effectively enhance model performance. This indicates that the complexity index is closely associated with semantic representations, vehicle kinematics, and contextual characteristics, and that hidden information can be effectively extracted using neural network structures.

\subsection{Ablation Studies} \label{sec:ablation}
Several experiments are conducted to justify the usability, stability, and robustness of the proposed Complexity-infused prediction model. 

\subsubsection{Comparison of LLM and MTurk Annotations}
The encoder trained on the MTurk annotations utilized a 32-hidden neuron structure and treated the output as a continuous variable, since the annotations were generated by averaging the workers’ responses. The performance of the encoders trained on LLM and MTurk data is shown in Table~\ref{tab:encoder_comparison}. It can be seen that the encoder using MTurk data performed slightly better than the one using LLM data. However, for the encoders trained on the combination of semantic, kinematic, and contextual features, the performance of both encoders was very similar.

\begin{table}[ht!]
\small
\begin{tabular}{llcc}
\hline
\begin{tabular}[c]{@{}c@{}}\textbf{Annotation}\\\textbf{Sources}\end{tabular}                                                                                             & \textbf{Input Features}              & \begin{tabular}[c]{@{}c@{}}\textbf{RMSE/CE} \\ \textbf{on Train} $\downarrow$\end{tabular} & \begin{tabular}[c]{@{}c@{}}\textbf{RMSE/CE} \\ \textbf{on Test} $\downarrow$\end{tabular} \\ \hline
\multirow{3}{*}{\begin{tabular}[c]{@{}l@{}}LLM\end{tabular}} & Semantic           & 1.10                                               & 1.10                                              \\
                                                                                                              & Semantic + kinematic & 1.05                                               & 1.06                                              \\
                                                                                                              & Sem. + Kin. + Cont.                  & 0.84                                               & 0.86                                              \\ \hline
\multirow{3}{*}{\begin{tabular}[c]{@{}l@{}}MTurk\end{tabular}}          & Semantic                    & 0.93                                                        & 0.94                                                       \\
                                                                                                              & Semantic + kinematic          & 0.91                                                        & 0.93                                                       \\
                                                                                                              & Sem. + Kin. + Cont.                           & 0.83                                                        & 0.88                                                       \\ \hline
\end{tabular}
\caption{Performance comparison of the complexity-infused encoder models using LLM data or MTurk data. The results are reported in terms of RMSE. Downward arrows next to the values indicate that lower RMSE is preferable.}
\label{tab:encoder_comparison}
\end{table}

Next, the crash prediction model was trained using the complexity-infused features derived from the MTurk data. The results of the crash likelihood prediction models trained on both LLM and MTurk data are shown in Table~\ref{tab:encoder_comparison_crash}. The model using complexity-infused features derived from LLM data consistently yielded higher accuracy than the one using MTurk data. This suggests that the LLM-generated complexity indexes may capture relevant information more effectively for crash likelihood prediction.

\begin{table}[ht!]
\footnotesize
\begin{tabular}{llcc}
\hline
\begin{tabular}[c]{@{}c@{}}\textbf{Annotation}\\\textbf{Sources}\end{tabular}             & \textbf{Input Features}                        &\begin{tabular}[c]{@{}c@{}}\textbf{Comp.-infused} \\ + Base $\uparrow$\end{tabular}   & \begin{tabular}[c]{@{}c@{}}\textbf{Comp.-infused} \\\textbf{alone} $\uparrow$\end{tabular} \\ \hline
\multirow{3}{*}{\begin{tabular}[c]{@{}l@{}}LLM\end{tabular}}            & Semantic                              & \textbf{78.35} & 65.31                                            \\
                                                                        & Semantic + kinematic                    & \textbf{86.32} & 67.85                                            \\
                                                                        & Sem. + Kin. + Cont.                                     & \textbf{90.15} & 78.49                                            \\ \hline
\multirow{3}{*}{\begin{tabular}[c]{@{}l@{}}MTurk\end{tabular}}          & Semantic                              & 71.01         & \textbf{67.72}                                             \\
                                                                        & Semantic + kinematic                    & 80.89         & \textbf{73.61}                                             \\
                                                                        & Sem. + Kin. + Cont.                                     & 86.60         & \textbf{80.06}                                             \\ \hline
\end{tabular}
\caption{Performance comparison of the crash prediction models using complexity-infused features from LLM data or MTurk data. The results are reported in terms of accuracy (\%). Upward arrows next to the values indicate that higher accuracy is preferable.}
\label{tab:encoder_comparison_crash}
\end{table}

It is also worth noting that the model using only the complexity-infused features from MTurk annotations outperformed the one based on LLM annotations. When considering the encoder performance (Table~\ref{tab:encoder_comparison}), the model trained solely on complexity-infused features from MTurk data showed better prediction capability than the ones from LLM data. However, when semantic, kinematic, and contextual features were added, the best-performing model was achieved using the LLM data. This implies that while the complexity-infused features derived from MTurk annotations capture important information effectively in isolation, the LLM-generated data provides a more robust integration of semantic, kinematic, and contextual features. As a result, the combination of LLM data with these additional features enhances the model’s overall performance.

\subsubsection{Comparisons of Other Encoders}
Similar to the experiments conducted in Section~\ref{sec:main_result}, the crash prediction model was trained using complexity-infused features from the second and third encoders, as presented in Table~\ref{tab:encoder_performance}. The results are shown in Table~\ref{tab:ablation_performance}. 

\begin{table}[ht!]
\small
\begin{tabular}{llc}
\hline
\textbf{Encoder}            & \textbf{Input Features}                        &\begin{tabular}[c]{@{}c@{}}\textbf{Base +} \\\textbf{Comp.-infused} $\uparrow$\end{tabular}   \\ \hline
\multirow{3}{*}{\begin{tabular}[c]{@{}l@{}}32 neurons, \\ 1-d cont. output\end{tabular}}            & Semantic                              & \textbf{78.35}                                              \\
                                                                        & Semantic + Kinematic                    & \textbf{86.32}                                              \\
                                                                        & Sem. + Kin. + Cont.                                     & \textbf{90.15}                                              \\ \hline
\multirow{3}{*}{\begin{tabular}[c]{@{}l@{}}16 neurons, \\ 1-d cont. output\end{tabular}}          & Semantic                              & 77.42                                                       \\
                                                                        & Semantic + Kinematic                    & 86.24                                                       \\
                                                                        & Sem. + Kin. + Cont.                                     & 89.56                                                       \\ \hline
\multirow{3}{*}{\begin{tabular}[c]{@{}l@{}}32 neurons, \\ 10-d cat. output\end{tabular}}          & Semantic                              & 75.07                                                       \\
                                                                        & Semantic + Kinematic                    & 83.59                                                       \\
                                                                        & Sem. + Kin. + Cont.                                     & 88.65                                                       \\ \hline
\end{tabular}
\caption{Performance comparison of the crash prediction models using complexity-infused features from different encoders. The results are reported in terms of accuracy (\%). Upward arrows next to the values indicate that higher accuracy is preferable.}
\label{tab:ablation_performance}
\end{table}

The models using different encoders consistently achieved higher accuracy as the number of input features increased. The encoder with 32 hidden neurons and a continuous output consistently delivered the highest accuracy, regardless of the input feature set. These results further validate the proposed encoder structure’s effectiveness in extracting hidden context from the existing semantic, kinematic, and contextual features through the complexity index.

\section{Conclusions and Limitations}
\label{sec:conclusion}
In this paper, we presented a two-stage framework for extracting hidden context from semantic, kinematic, and contextual features, and for predicting crash density by incorporating these hidden context features into the original feature sets. This approach addresses the challenges of identifying and understanding the key factors associated with crash prevalence in real-world naturalistic driving environments. To our knowledge, this is the first model to integrate all scene-related and driving-related features and link them to a complexity index obtained from the LLM. Our experiments revealed that the framework can accurately predict crash density, achieving an accuracy of $90.15\%$. The data generated by LLM provided better predictive capability than that obtained from human annotation via MTurk. Different encoder structures consistently outperformed models without complexity-infused features. This solution has the potential to inform the development of advanced driver assistance systems, and driver monitoring systems to improve safety in manually driven and automated vehicles. The work can lead to enhanced human-AI collaboration that more effectively supports drivers through complex environmental changes. Furthermore, insights from this framework have the potential to support roadway design by empowering highway engineering departments with the data needed to identify and mitigate risk factors in crash hotspots. If embraced through infrastructure changes these results would thus contribute to a key proactive traffic safety intervention that aligns with US DOT's Safe-Systems Approach.

In the future, to further improve the generalizability of the model, we aim to expand the dataset by incorporating more diverse driving environments and a broader range of driving conditions, such as adverse weather and complex urban settings. Furthermore, we intend to experiment with more advanced encoder architectures to refine the extraction of hidden context features. Finally, we plan to investigate the integration of real-time data, enabling dynamic crash likelihood predictions for real-world applications. 

\section{Acknowledgment}
Data for this study were drawn from work supported by the Advanced Vehicle Technologies (AVT) Consortium at MIT (\url{http://agelab.mit.edu/avt}). The views and conclusions expressed are those of the authors and have not been sponsored, approved, or endorsed by the Consortium or Foundation.

This paper is funded partially by the New England University Transportation Center (NEUTC). NEUTC is funded by a grant from the U.S. Department of Transportation’s University Transportation Centers Program (69A3552348301). However, the U.S. Government assumes no liability for the contents or use thereof.

{
    \small
    \bibliographystyle{ieeenat_fullname}
    \bibliography{main}
}

\clearpage
\setcounter{page}{1}
\maketitlesupplementary

\section{LLM Implementation Details} \label{sup:LLM}
The contextual features and the complexity index feature mentioned in~\cref{cont} and~\cref{comp} were generated using the \texttt{GPT-4o-2024-08-06} model \footnote{https://platform.openai.com/docs/models/gpt-4o}. Batch API \footnote{https://platform.openai.com/docs/guides/batch} was used for faster processing.

\section{Annotation Comparisons} \label{sup:annoComp}
Table~\ref{tab:comp} presents the descriptive statistics for the comparison between AI-generated and human-generated annotated complexity indices, as discussed in \cref{comp}. This comparison is based on the entire dataset without matching. It can be observed that the overall distribution of the complexity index annotated by AI and humans is notably similar. 

\begin{table}[h]
    \small
    \centering
    \begin{tabular}{lccccccc}
        \toprule
        \textbf{Method} & \textbf{M (SD)} & \textbf{Min} & \textbf{25\%} & \textbf{50\%} & \textbf{75\%} & \textbf{Max} \\
        \midrule
        AI    & 4.2 (1.3) & 1 & 3 & 4 & 5 & 8 \\
        Human & 4.5 (1.2) & 1.7 & 3.6 & 4.5 & 5.2 & 8.2 \\
        \bottomrule
    \end{tabular}
    \caption{The descriptive statistics of the complexity index from AI-generated and human-generated annotations.}
    \label{tab:comp}
\end{table}

When looking at the difference in complexity indices between AI and human annotations (diff $=$ complexity index of AI $-$ complexity index of human), the difference has a mean of $-0.25$, a standard deviation of $1.07$, a minimum of $-4.5$, a 25th percentile of $-1$, a median of $-0.3$, a 75th percentile of $0.4$, and a maximum of $3.2$. It can be observed that the differences are centered around zero with AI annotations being slightly smaller than human annotations on average.

\section{Model Structure} \label{sup:structure}
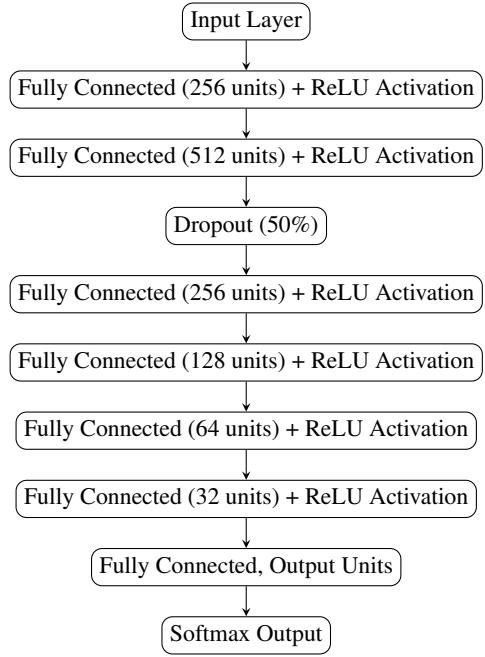
\begin{figure}[h]
\begin{tikzpicture}[node distance=.4cm and .4cm, >=stealth, scale=0.7, every node/.style={scale=0.9}]
    \node[draw, rounded corners] (input) {Input Layer};

    \node[draw, rounded corners, below=of input] (fc1) {Fully Connected (256 units) + ReLU Activation};
    
    \node[draw, rounded corners, below=of fc1] (fc2) {Fully Connected (512 units) + ReLU Activation};
    
    \node[draw, rounded corners, below=of fc2] (dropout) {Dropout (50\%)};

    \node[draw, rounded corners, below=of dropout] (fc3) {Fully Connected (256 units) + ReLU Activation};
    
    \node[draw, rounded corners, below=of fc3] (fc4) {Fully Connected (128 units) + ReLU Activation};
    
    \node[draw, rounded corners, below=of fc4] (fc5) {Fully Connected (64 units) + ReLU Activation};
    
    \node[draw, rounded corners, below=of fc5] (fc6) {Fully Connected (32 units) + ReLU Activation};

    \node[draw, rounded corners, below=of fc6] (fc7) {Fully Connected, Output Units};
    \node[draw, rounded corners, below=of fc7] (softmax) {Softmax Output};

    \draw[->] (input) -- (fc1);
    \draw[->] (fc1) -- (fc2);
    \draw[->] (fc2) -- (dropout);
    \draw[->] (dropout) -- (fc3);
    \draw[->] (fc3) -- (fc4);
    \draw[->] (fc4) -- (fc5);
    \draw[->] (fc5) -- (fc6);
    \draw[->] (fc6) -- (fc7);
    \draw[->] (fc7) -- (softmax);
\end{tikzpicture}
\caption{Visualization of the crash density prediction neural network model architecture.}
\label{fig:modelStructure}
\end{figure}

The neural network model structure mentioned in~\cref{crashDensity} is shown in Fig.~\ref{fig:modelStructure}.

\section{Model Training Details}
\subsection{Encoder Training Details} \label{sup:encoderTrainingDetails}
Regarding the training configurations of the encoders mentioned in~\cref{sec:encoder}, different setups were applied based on the input features and the output data types. The model was trained for $1000$ epochs using Stochastic Gradient Descent (SGD) as the optimizer with a momentum value of $0.9$. The learning rate varied depending on the feature and data type:
\begin{itemize}
    \item Semantic features only and continuous data type: Learning rate of $0.0005$.
    \item Semantic features only and categorical data type: Learning rate of $0.005$.
    \item Semantic and driving features and continuous data type: Learning rate of $0.0003$.
    \item Semantic and driving features and categorical data type: Learning rate of $0.001$.
    \item All available features (semantic, driving, and contextual) and continuous data type: Learning rate of $0.0005$.
    \item All available features (semantic, driving, and contextual) and categorical data type: Learning rate of $0.001$.
\end{itemize}
A Cosine Annealing learning rate scheduler was used to adjust the learning rate over the training epochs.

\subsection{Prediction Model Training Details} \label{sup:predictionTrainingDetails}
Regarding the training configurations of the prediction models mentioned in \cref{sec:main_result}, all models were trained using SGD with a momentum value of $0.9$. The learning rate was set to $0.001$, with a Cosine Annealing scheduler. The total number of epochs was $2000$.


\end{document}